\documentclass[11pt]{article}
\usepackage[margin=1in]{geometry}
\usepackage{booktabs}
\usepackage{amsmath}
\usepackage{natbib}
\usepackage[hidelinks]{hyperref}
\usepackage{microtype}

\title{Memory That Looks Forward: A Zero-Inference Prospective Term\\for Personal Memory Retrieval}
\author{Jonathan Groff\\ \small Independent Researcher\thanks{The author is a doctoral student in Strategic Leadership at Liberty University. This work was conceived and conducted independently, in the author's personal capacity, and is unaffiliated with, and does not represent the views of, any institution or employer.}}
\date{}

\begin{document}
\maketitle

\begin{abstract}
Retrieval over a personal memory store is retrospective: it surfaces what resembles the query, and it is blind to what the user has committed to do. We describe a prospective term for memory retrieval that costs no inference at query time. Commitments are held in an explicit ledger as dated or trigger-conditioned entries; memory items linked to a firing entry receive a salience boost, blended multiplicatively into embedding-based retrieval so that relevance remains sovereign. On a synthetic prospective-memory task set modeled on TriggerBench's published structure (48 blind-authored dialogues, 175 tasks), the term raised recall@5 on the hard stratum from 0.000 to 0.955 at the default blend weight and to 1.000 under a floor variant, with zero false boosts across 53 resolved-commitment tasks. Blind authorship also produced a scope finding: only 17--29\% of naturally phrased commitment--trigger pairs defeat embedding similarity, so the term matters on a real minority of cases and must do no harm on the rest, which it does not. We position precomputed commitment linkage as the always-on floor of a layered design whose expansion layer is query-time prospection. Results are preliminary: the evaluation set is author-constructed, and evaluation on TriggerBench proper is committed follow-up work once its data is released.
\end{abstract}

\section{Introduction}
A person managing their life through a personal knowledge store mentions a commitment once, in passing, the way people actually speak: an invoice due on the fifteenth, a valve that needs replacing before guests arrive. Twenty conversations later a situation arrives that makes the commitment operative, and the situation shares almost no vocabulary with the original mention. Embedding-based retrieval, the default memory mechanism of current assistant systems, misses it. The failure is structural. Dense retrieval ranks by resemblance to the query, and the trigger does not resemble the constraint \citep{zhang2026triggerbench}.

Human memory does not work this way. Prospective memory, remembering to act when a future condition arrives, is a distinct faculty from recalling the past, and it is the faculty that makes a second brain more than an archive. A human is always thinking about what is next. The machine should also be thinking about what is next, along with the human, and help guide collaborative thinking. That stance motivates the design here: this component is the first stage of an architecture aimed at proactive reasoning and anticipatory execution, in which viewing the horizon acts as a trigger response to stimuli and supplies first-pass information that guides both the user and the machine toward better outcomes.

The design question is where the anticipation should be computed. Prospection-Guided Retrieval \citep{chopra2026pgr} generates imagined futures at query time and retrieves against them, recovering memories that lie far from the query in embedding space, at the cost of language-model inference on every query. We take the opposite corner of the design space: a \textbf{zero-inference prospective term}. Commitments live in an explicit ledger as first-class memory objects, each dated or trigger-conditioned; memory items linked to a firing entry receive a precomputed salience boost; the boost blends multiplicatively into cosine retrieval so a boosted item can rise only among candidates that are already relevant. No model runs at query time.

This paper contributes: (1) the design of the prospective term and the salience model it lives in, including the blending rule and the safety principle that salience may rank memories and may never prune them; (2) a synthetic prospective-memory evaluation built on TriggerBench's published task structure under a blind-authorship protocol, with results across difficulty strata, a boost-weight sweep, and resolution-tracking negatives; (3) a scope finding from the blind authorship itself, quantifying how often natural phrasing actually defeats embedding retrieval; and (4) a measured boundary of the multiplicative blend, with a floor variant that removes it.

\section{Related work}
\paragraph{Prospective memory for language models.} TriggerBench \citep{zhang2026triggerbench} establishes the task: a latent constraint stated early, distractor turns, then a trigger that makes the constraint operative, with negative variants in which the constraint is resolved and the correct behavior is silence. Their diagnosis of retrieval failure, that triggers lack lexical and semantic overlap with latent constraints, is the exact phenomenon our term targets; their RAG baseline reaches 51.62\% PM accuracy. Prospection-Guided Retrieval \citep{chopra2026pgr} attacks the same gap from the query side, expanding a goal into a short tree of plausible next steps and retrieving against those steps, with large recall gains and query-time inference cost.

\paragraph{Salience-inspired retrieval.} Generative agents score memories by recency, relevance, and importance \citep{park2023generative}; ACT-R models activation as frequency- and recency-decayed base level plus spreading activation \citep{anderson2004integrated}; HippoRAG applies graph centrality to memory \citep{gutierrez2024hipporag}. We treat this literature as design inspiration and build no empirical claim on it: an adversarial verification pass we ran against primary sources could not confirm the load-bearing claims we had attributed to it, and the honest posture is to say so. One adjacent finding we do rely on: memory selection optimized for utility systematically forgets constraint-type memories \citep{lin2026safety}, which is why salience in our system is a ranking signal and never a deletion criterion.

\paragraph{Temporal memory substrates.} Zep's temporal knowledge graph \citep{rasmussen2025zep} supplies the substrate conventions our store follows, in particular invalidation instead of deletion, which is what allows a resolved commitment to stop boosting without destroying the record that it existed.

\section{Design}
\paragraph{Setting.} A single-user, human-curated markdown knowledge vault. Retrieval is hybrid: a structural index and link graph, plus local embeddings (bge-micro-v2, 384 dimensions) with cosine ranking. A standing constraint of the system is that no language model runs at query time; retrieval is deterministic arithmetic over precomputed vectors.

\paragraph{The commitments ledger.} Prospective memory is explicit. A ledger holds open commitments in two forms: dated entries (``submit the registration packet by 15 August'') and trigger-conditioned entries (``when the vendor quote arrives, reconcile it against the June estimate''). Entries name the action, the condition or date, and links to the memory items they concern. Entries are closed or resolved when satisfied, and resolution is recorded, never deleted. All examples in this paper are synthetic.

\paragraph{The salience model.} Each memory item carries a standing salience score, computed offline from five components: intrinsic importance (weight 0.30), usage activation with slow decay (0.25), link-graph centrality (0.20), \textbf{prospective linkage (0.15)}, and recency (0.10, deliberately the smallest voice, so that old but recurring material is not eclipsed by the merely new). The prospective component assigns its boost to items linked from open ledger entries, with a larger boost when an entry is firing: its date has arrived or its trigger condition matches the current context. The weights are declared hypotheses, not fitted parameters.

\subsection{The salience computation, precisely}

Each memory item $i$ receives $s_i = 0.30\,I_i + 0.25\,A_i + 0.20\,C_i + 0.15\,P_i + 0.10\,R_i$, with every component normalized to $[0,1]$ across the corpus before weighting (division by the corpus maximum). Intrinsic importance $I_i$ takes values 1.0, 0.6, or 0.3 from an explicit per-item annotation, with defaults inferred from item type (indexes and reference anchors 1.0; dated ephemera 0.3; all else 0.6). Activation $A_i$ is an ACT-R-style base level over the system's append-only operations log: the sum over recorded mentions of $(d+1)^{-0.5}$, where $d$ is days since the mention. Centrality $C_i$ is PageRank (damping 0.85) over the store's link graph. Prospective linkage $P_i$ is 1.0 for items linked from an open ledger entry, 0.6 for items sharing a project with one, else 0. Recency $R_i$ is $(d+1)^{-0.5}$ in days since last modification.

\paragraph{What was active in the reported experiments.} The evaluation isolates the prospective term. For each candidate turn, the blend uses $\mathrm{score} = \cos 	imes (1 + W \cdot b)$ with $b \in \{0, 1\}$: $b = 1$ exactly when the turn is linked from a ledger entry that is firing (its date arrived or its condition matched) and unresolved, else $b = 0$. The other four components were deliberately inactive, so the reported numbers measure the prospective term alone, uncontaminated by importance, activation, centrality, or recency effects; this is also why the oracle results read as an upper bound on that term specifically, not on the full blend.

\paragraph{Blending.} Salience enters retrieval multiplicatively: $\mathrm{score} = \cos \times (1 + W \cdot s)$, default $W = 0.3$. We adopted this form after observing the additive alternative fail: adding salience to cosine let important but off-topic items crowd into results for unrelated queries. Multiplication keeps relevance sovereign. A boosted item rises among comparably relevant candidates and cannot displace them from nowhere. The risk of a prospective term is a narrowing of focus; the counter is that the aperture stays wide from the user's input, because the term is one voice in the blend and the query remains the ranker.

\paragraph{Safety principle.} Salience ranks; it never prunes. Importance-weighted deletion demonstrably forgets constraint-type memories first \citep{lin2026safety}, and a deleted memory cannot re-emerge on its own. Pruning in the surrounding system is structural (age and lift status) and is out of scope here.

\section{Evaluation}
\subsection{Task set construction}
We model TriggerBench's published structure without using its data, which is unreleased at the time of writing: each blueprint is a multi-turn dialogue with a constraint stated naturally (never as an explicit reminder request), 15--30 distractor turns, and a trigger whose surface wording does not resemble the constraint. Variants: positive-clean, positive-overloaded (trigger embedded among concurrent requests), and negative-clean (an intermediate turn resolves the constraint; correct behavior is no boost). Dimensions: temporal grounding, state tracking, and logical adherence.

\paragraph{Blind authorship.} A pilot of five blueprints, whose wording had been iterated against the scoring embedder, taught us that tuned difficulty is manufactured difficulty. The scaled set of 48 blueprints was therefore authored without consulting the embedder and measured exactly once. The primary difficulty gate is relative: a blueprint is \textbf{hard} if the constraint's baseline cosine rank falls in the bottom half of its dialogue's turns, and \textbf{easy} otherwise. Nothing was discarded; easy blueprints form a retained stratum. One batch-level guidance revision (avoid shared named entities; make triggers situational instead of referential) was applied between batches and is disclosed; no per-item wording was ever revised against a measurement. The pilot's five blueprints are reported as a separate tuned stratum and never as headline numbers. An earlier absolute gate (entity-masked cosine below 0.50) is reported alongside for every blueprint; it would have discarded 85\% of blind-authored blueprints, which retroactively quantifies how much author leakage had shaped the pilot.

\paragraph{Integrity checks.} An independent consistency audit of all 48 blueprints (date coherence, resolution validity, self-contradiction) found one blocking defect, which was repaired without touching the constraint or trigger. All dialogues were LLM-authored under the protocol above; the author, who wrote none of them, spot-checked a stratified sample of six for naturalness and judged them adequate for the experiment's purpose. We report that check for what it is: sample-level and single-reviewer.

\subsection{Protocol and metrics}
The evaluation isolates retrieval. At the trigger turn, all prior turns are candidates; the baseline ranks them by cosine similarity to the trigger; the blend applies the prospective boost for firing ledger entries, built here from oracle annotations (the constraint annotations themselves), which makes the results an upper bound on the scorer under perfect ledger extraction. Metrics: recall@1, recall@5, and MRR of the constraint turn on positives; on negatives, the false-boost rate, the fraction of resolved constraints that a blend lifts into the top five when the baseline does not. A held-out paraphrase set (16 triggers reworded by an author who never saw the matcher's rules) measures the condition matcher's brittleness. The full set is 175 tasks. Everything is deterministic; the current date is whatever the trigger asserts.

\subsection{Results}
\begin{table}[t]
\centering
\small
\begin{tabular}{lccc@{\hskip 1.5em}ccc}
\toprule
& \multicolumn{3}{c}{Hard stratum ($n=22$)} & \multicolumn{3}{c}{Easy stratum ($n=74$)} \\
\cmidrule(r){2-4}\cmidrule(l){5-7}
Scorer & R@1 & R@5 & MRR & R@1 & R@5 & MRR \\
\midrule
baseline (cosine)      & 0.000 & 0.000 & 0.077 & 0.122 & 0.554 & 0.312 \\
$\times(1+0.1s)$       & 0.000 & 0.318 & 0.152 & 0.527 & 0.946 & 0.688 \\
$\times(1+0.3s)$       & 0.545 & 0.955 & 0.701 & 0.905 & 1.000 & 0.953 \\
$\times(1+0.5s)$       & 0.909 & 1.000 & 0.955 & 0.973 & 1.000 & 0.986 \\
$\times(1+1.0s)$       & 1.000 & 1.000 & 1.000 & 0.973 & 1.000 & 0.986 \\
additive floor         & 1.000 & 1.000 & 1.000 & 0.973 & 1.000 & 0.986 \\
\bottomrule
\end{tabular}
\caption{Positive-task retrieval of the constraint turn, blind-authored strata. The tuned pilot stratum is reported in supplementary material and is excluded from headline results.}
\label{tab:main}
\end{table}

\paragraph{Hard stratum} (22 positive tasks): the baseline never places the constraint in the top five (recall@5 0.000, MRR 0.077). The default blend reaches recall@5 0.955 and MRR 0.701. Raising $W$ to 0.5 gives 1.000 and 0.955; $W \ge 1.0$, and a floor variant that promotes firing entries into the candidate set outright, reach 1.000 on both (Table~\ref{tab:main}).

\paragraph{Easy stratum} (74 positive tasks): the baseline already performs (recall@5 0.554), and the blend improves it (1.000 at $W=0.3$) with no degradation observed at any weight. The component helps where needed and does no harm where it is not.

\paragraph{Negatives} (53 tasks across strata): zero false boosts at every weight. A diagnostic confirms the result is earned, not vacuous: ignoring resolution, the matcher would have fired on all 53.

\paragraph{The multiplicative ceiling, quantified.} The pilot observed that a small multiplier cannot rescue a deeply buried constraint. On fired hard positives, the multiplier needed for rank one has median 0.29 and maximum 0.56; $W=0.3$ suffices for 55\% of top-1 placements, $W=0.5$ for 91\%, $W=1.0$ for all. The floor variant removes the ceiling by construction. The practical reading: recall@5 is forgiving of a timid weight, top-1 is not, and deployments that need top-1 should either raise $W$ for firing entries or floor them.

\paragraph{Matcher brittleness.} On held-out paraphrases the keyword-and-date matcher fired on 13 of 16 (81\%). Unfired tasks fall back to baseline ranking, and hard-stratum paraphrase recall@5 drops to 0.818 accordingly. The condition matcher, not the scorer, is the component's weakest surface.

\paragraph{The scope finding.} Blind authorship doubles as a measurement of the phenomenon's base rate: 17\% of batch-1 and 29\% of batch-2 blueprints were hard under the relative gate. Triggers that engage a commitment's situation usually do share vocabulary with the constraint. The hard case this term targets is real and it is the minority. We consider this the evaluation's most transferable result: a prospective term should be judged on the hard minority and on doing no harm to the easy majority, and claims that omit the base rate overstate the stakes.

\section{Discussion}
\paragraph{Notes and intention.} Precomputed links and imagined futures are not rivals. Precomputed linkage is reference: it can only surface what already exists in the notes. Query-time prospection expands: it questions beyond the artifacts and keeps the aperture open. You need both. You need good notes, and you need intention, and the imagined futures are the intention. The engineering consequence is a layered design: the zero-inference term as the always-on floor, catching every commitment the user made explicit, and simulation as the escalation for queries where the future must be guessed instead of consulted. The floor costs nothing per query; the escalation earns its inference budget only where the floor comes up empty.

\paragraph{What the term is for.} A purely historical note store is a solved problem, and a sophisticated search over it is still a search engine. Layering an anticipatory system over the historical archive moves the artifact closer to how a person's memory serves them, and a system modeled on that serves its user more than resemblance ranking can. The scope finding disciplines this ambition: most of the time, resemblance is enough, and the anticipatory layer justifies itself on the cases where it is not, which are exactly the cases users experience as the system forgetting what they told it.

\paragraph{Failure mode.} The component's ceiling is the ledger. If the ledger accumulates irrelevant entries or false direction, the boost degrades to noise; quality over time is the failure condition, and constant correction is required. The bound is the user's own input discipline, and we consider that an honest property of the design instead of a removable defect: the term amplifies expressed intention and cannot supply intention that was never expressed.

\section{Limitations}
The evaluation is synthetic and author-constructed; TriggerBench proper is committed follow-up work once its data is released, and until then all numbers here are preliminary. The oracle ledger makes results an upper bound: real deployments must extract ledger entries at ingest, and the extraction tier is unbuilt. The hard stratum is small (11 blueprints, 22 positive tasks); we report direction, not magnitude. The relative gate was adopted after the absolute gate's failure was observed, a disclosed post-hoc change mitigated by reporting both gates for every blueprint. Dialogues were authored by a language model whose phrasing habits may correlate with the embedder's geometry. The system under test is a single user's; nothing here claims generality across users or stores. Finally, the salience weights are hypotheses under live observation, not fitted values.

\section{Conclusion}
A memory system that only looks backward answers questions; one that also looks forward keeps commitments. We showed that a prospective term built from an explicit ledger, precomputed links, and a multiplicative blend recovers commitments that embedding retrieval misses entirely, at zero query-time inference cost and zero measured false alarms, while leaving the easy majority of cases untouched. The boundary of the approach is equally clear: it retrieves only intentions the user expressed, its blend has a measurable ceiling that a floor removes, and its evaluation awaits confirmation on the benchmark whose structure it borrows. New information is a request for a new decision. A memory that surfaces the right commitment at the right moment is how a system makes that request on time.

\section*{Acknowledgments and AI disclosure}
The memory system described here was co-constructed, and this paper drafted, with a large language model assistant (Anthropic Claude); the evaluation dialogues were LLM-authored under the blind protocol of Section~4.1; all design decisions, the interview material the paper is built from, and the naturalness spot-check are the author's. The plural voice throughout is deliberate: the system was built and studied by the author working with the assistant, and ``we'' reflects that collaboration. Authorship, responsibility, and every claim in the paper are the author's alone.

\paragraph{Data availability.} The synthetic task set (blueprints and all 175 instantiated tasks), the evaluation harness, and per-task results contain no personal data and are released at \url{https://github.com/Groffitti/memory-that-looks-forward}. Results will be reiterated on TriggerBench proper once its dataset is publicly available.

\end{document}